# Towards Efficient Unconstrained Palmprint Recognition via Deep Distillation Hashing

Huikai Shao, *Student Member IEEE*, Dexing Zhong, *Member IEEE*, Xuefeng Du

*Abstract*—**Deep palmprint recognition has become an emerging issue with great potential for personal authentication on handheld and wearable consumer devices. Previous studies of palmprint recognition are mainly based on constrained datasets collected by dedicated devices in controlled environments, which has to reduce the flexibility and convenience. In addition, general deep palmprint recognition algorithms are often too heavy to meet the real-time requirements of embedded system. In this paper, a new palmprint benchmark is established, which consists of more than 20,000 images collected by 5 brands of smart phones in an unconstrained manner. Each image has been manually labeled with 14 key points for region of interest (ROI) extraction. Further, the approach called Deep Distillation Hashing (DDH) is proposed as benchmark for efficient deep palmprint recognition. Palmprint images are converted to binary codes to improve the efficiency of feature matching. Derived from knowledge distillation, novel distillation loss functions are constructed to compress deep model to further improve the efficiency of feature extraction on light network. Comprehensive experiments are conducted on both constrained and unconstrained palmprint databases. Using DDH, the accuracy of palmprint identification can be increased by up to 11.37%, and the Equal Error Rate (EER) of palmprint verification can be reduced by up to 3.11%. The results indicate the feasibility of our database, and DDH can outperform other baselines to achieve the state-of-the-art performance. The collected dataset and related source codes are publicly available at *http://gr.xjtu.edu.cn/web/bell/resource*.**

## I. INTRODUCTION

In digital society, personal authentication is becoming more and more important for information security. However, some traditional methods for personal authentication, such as keys and passwords, have some drawbacks. In recent years, biometrics has been regarded as one of the most important and effective solutions for personal authentication. Biometrics is an efficient technique that uses the physiological or behavioral properties of human body for identity recognition [1]. Generally, there are two recognition modes for a biometrics system, *i.e.* identification and verification [2]. Identification is a one-to-many comparison, which answers the question of "who the person is?". Verification is a one-to-one comparison, which answers the question of "whether the person is whom he claims to be". As one of the emerging biometrics technique, palmprint recognition has received wide attention from researchers [3]. To date, many effective palmprint recognition methods have been proposed, such as Robust Line Orientation Code (RLOC) [4], Linear Programming (LP) formulation [5], Discriminative and Robust Competitive Code (DRCC) [6], and Deep learning-based methods [7]. Zhao and Zhang [8] also proposed salient and discriminative descriptor learning method (SDDLM) for palmprint recognition in general scenarios, such as contact-based and contactless recognition. These methods have achieved promising performance on some public palmprint databases, such as PolyU multispectral dataset [9].

However, current palmprint recognition systems still have some shortcomings. Firstly, the requirements of dedicated palmprint acquisition devices limit the convenience of palmprint recognition. For example, in order to avoid external illumination interference, most available palmprint databases were collected in enclosed space with additional lights in constrained manners. Volunteers need to put their hands on a certain device, and in some cases they have to grip a handle to obtain stable images, which inevitably reduces the user-friendliness. Secondly, though deep learning-based methods can have the merit of high accuracy, they still have the shortcomings of high complexity and low efficiency, which rarely meets the requirements of practical applications. Therefore, efficient unconstrained palmprint recognition has been an urgent issue and re-awakened interest in the use for palmprint [10].

In order to well address above problems, in this paper, we establish a new unconstrained palmprint database and propose Deep Distillation Hashing (DDH) as benchmark for efficient deep palmprint recognition. Firstly, using 5 different mobile phones, a large palmprint database is collected named as Xi'an Jiaotong University Unconstrained Palmprint (XJTU-UP) database. In this database, more than 20,000 palmprint images are captured from 100 individuals in different environments in

This work is supported by the National Natural Science Foundation of China (No. 61105021), Natural Science Foundation of Zhejiang Province (No. LGF19F030002), and Natural Science Foundation of Shaanxi Province (No. 2020JM-073).

H. Shao is with the School of Electronic and Information Engineering, Xi'an Jiaotong University, Xi'an, Shaanxi 710049, China (e-mail: shaohuikai@stu.xjtu.edu.cn).

D. Zhong is with the School of Electronic and Information Engineering, Xi'an Jiaotong University, Xi'an, Shaanxi 710049, China, and State Key Lab. for Novel Software Technology, Nanjing University, Nanjing, 210093, P.R. China (e-mail: bell@xjtu.edu.cn).

X. Du is with the School of Electronic , Xi'an Jiaotong University, Xi'an, Shaanxi 710049, China (dxfsxl@163.com).

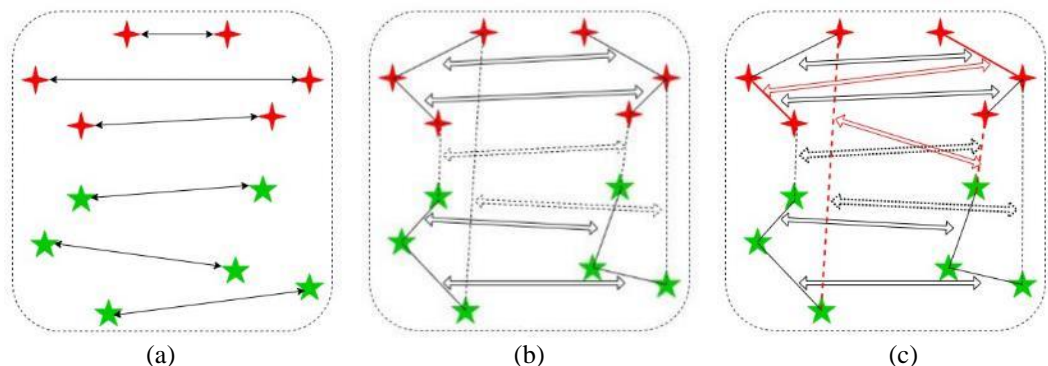

**Fig. 1.** Illustration of different distillation losses. In each sub-figure, the left shows features extracted by teacher network, and the right shows the ones by student network. Different colors and graphics represent different categories. KD is aimed to make the features extracted by teacher and student networks as similar as possible. In (a), the distance between features of the same samples extracted by teacher and student networks is minimized (point to point). In (b), the solid line represents the distance of genuines, the dashed line represents the distance of imposters, and the double arrow indicates optimization direction. The distance of each sample matching pair in teacher and student networks is adopted as distillation loss (distance to distance). On the basis of (b), (c) adds a new constraint. It makes the maximum distance of student genuines smaller than the minimum distance of teacher genuines (red solid line), and the minimum distance of student imposters larger than the maximum distance of teacher imposters (red dashed line). Best viewed in color.

an unconstrained manner using two kinds of illuminations. Due to the complex background, traditional region of interest (ROI) extraction algorithms based on the contours and valleys of hands [11] are not available. In order to promote the research of ROI extraction on XJTU-UP database and extract stable ROIs, each palmprint image is manually marked with 14 key points on the contour of palm, including the valley points between fingers, which are also publicly available to the research community. Based on these key points, a method for manually extracting ROI is given below. The details of XJTU-UP database can be found in Section III.

Secondly, DDH is proposed for efficient palmprint recognition, which can convert the palmprint images into binary codes to improve the effectiveness of feature matching using light network. The mechanism of knowledge distillation (KD) is introduced to reduce the complexity of model, where VGG-16-based Deep Hashing Network (DHN) is constructed as a teacher network and a light model with only two convolutional layers and three fully connected layers is selected as the student network. KD is aimed to make the features extracted by the teacher and student networks as similar as possible to help learn student network. In our previous work [12], the Euclidean distance between features of the same samples extracted by teacher and student networks is directly minimized. In [13], Yu *et al*. proposed a novel distillation loss to improve the performance of metric learning in student network, which constrained the distance between sample matching pairs in teacher and student networks, including genuine matchings and imposter matchings. Inspired by this idea, a new distillation loss is proposed in this paper, which wants to make the maximum distance of student genuine matchings smaller than the minimum distance of teacher genuine matchings, and the minimum distance of student imposter matchings larger than the maximum distance of teacher imposter matchings. A graphical illustration of these distillation losses is shown in Fig. 1. In general, complex teacher network can extract obvious features with small intra-class distances and large inter-class distances. Therefore, the distance of genuine and imposter matchings in teacher network can be adopted to guide the student network to obtain similar features. As a results, the student network can achieve higher recognition accuracy and efficiency. The details of DDH can be found in Section IV.

The contributions can be summarized as follows:

1. We publish a new palmprint database named as XJTU-UP. To the best of our knowledge, XJTU-UP is the largest unconstrained palmprint database currently published. Particularly, using 5 mobile phone, XJTU-UP database is collected in complex backgrounds in an unconstrained manner. And it is also a database across different devices including up to 10 sub-datasets. Owing to the characteristics of XJTU-UP database, it can be used for multiple research tasks, such as palmprint identification and verification, cross-domain palmprint recognition, and open-set palmprint recognition. In order to conveniently extract ROI, 14 key points are manually marked and publicly available to the research community too.

2. DDH is proposed as benchmark for efficient palmprint recognition based on DHN and knowledge distillation algorithms. Palmprint images are converted into binary codes, which can reduce storage space and matching time. Based on knowledge distillation, novel effective distillation loss functions are constructed to reduce the difference of feature distributions between teacher network and light student network. As a result, the student network can achieve efficient performance under the guidance of teacher network.

3. Extensive experiments of palmprint identification and verification are conducted on XJTU-UP database and other benchmark palmprint databases. The results indicate that the proposed DDH is more robust and efficient in dealing with palmprint identification and verification as compared with the other state-of-the-art baselines. Further, compared with student baseline, the accuracy of palmprint identification is increased by up to 11.37% and the Equal Error Rate (EER) of palmprint verification is reduced by up to 3.11% using DDH.

Compared with our previous work in [12], we have made many significant improvements. Firstly, a novel and efficient distillation loss function is constructed to improve the communication of teacher network to student network, which can drive the student features closer to the teacher features. Results show that DDH can improve the accuracy of palmprint recognition by up to 11.37%. Secondly, a new task of palmprint identification with impressive results are performed to further evaluate the effectiveness of our modified algorithm. Thirdly, we add a more comprehensive analysis for XJTU-UP database, and compare it with the currently available palmprint databases in a table. Fourthly, two other benchmarks, Tongji contactless [14] and PolyU multispectral palmprint databases, are also introduced to carry out palmprint identification and verification, which also demonstrates the robustness of DDH. In addition, more adequate analyses and comparisons with the state-of-the-art baselines are conducted to evaluate the superiority of our algorithms.

The rest of this paper is organized as follows: Section 2 introduces some related works. Section 3 describes our XJTU-UP database in detail. Section 4 explains our proposed DDH. The experiments and results are presented in section 5. The evaluation and analyses of results are given in section 6. Section 7 concludes the paper.

## II. RELATED WORK

### A. Palmprint Recognition

Typically, a palmprint recognition pipeline consists of the steps of image acquisition, preprocessing, feature extraction and matching [11]. For image acquisition, different devices have been designed, and a lot of popular databases are collected and publicly available. Zhang et al. [15] captured 7,752 gray-scale palmprint images in a contacted way with a fixator. They also constructed the PolyU multispectral dataset containing 24,000 palmprint images under four spectral bands [9]. Hao et al. created a contactless palmprint dataset called as CASIA database, which contains 5,502 images captured from 624 hands in two sessions [16] [17]. In addition, the Indian Institute of Technology Delhi database (IIT-D) touchlessly collected 2,601 hand images from 460 individuals within an enclosed device [18]. In 2017, Zhang et al. [14] from Tongji University established a larger contactless palmprint database containing 12,000 palmprint images captured from 600 hands under the constrained environment. Matkowski et al. [19] collected 7,881 palm images of 2,035 hands from the Internet to establish a new palmprint database named NTU Palmprints (NTU-PI-v1). Besides, there are also some databases acquired by mobile devices. Choraś and Kozik et al. [20] used a smartphone to collect 252 palmprint images from 84 individuals. Jia et al. [21] used 2 smartphone cameras and a compact camera to collect 12,000 palmprint images under two different illumination conditions for cross device recognition. Even though there are a lot of mobile phone-based palmprint databases, most of them are not public to research communities. To our best knowledge, only Adrian-Stefan Ungureanu et al.

[22] publicized their NUIG_Palm1 dataset collected by 5 smartphone cameras.

According to different feature extraction mechanisms, existing palmprint recognition methods can be divided into several categories, such as structure-based and texture-based methods [23]. Kong et al. [24] extracted orientation information from palm lines and proposed Competitive Coding. Guo et al. [25] proposed binary orientation co-occurrence vector (BOCV) to represent multiple orientations for a local region. After that, Zhang et al. [26] incorporated fragile bits information in the code maps from different palms and proposed extended binary orientation co-occurrence vector (E-BOCV). Dai et al. [27] designed a ridge-based palmprint matching system. It quantitatively studied the statistical characteristics of the main features in the palmprint. Minaee and Abdolrashidi [28] proposed Discrete cosine transform (DCT)-based features in parallel with wavelet-based ones for palmprint recognition. They used Principal Component Analysis (PCA) to reduce the dimensionality of features and majority voting algorithm to perform classification. Xu et al. [29] proposed a sparse representation method for bimodal palmprint recognition. They accomplished the feature level fusion by combining the samples of different biometric traits into a vector and used the approximate representation to classify the test samples. Li and Kim [30] proposed Local Microstructure Tetra Pattern (LMTrP) for palmprint recognition, which took advantage of local descriptors' direction as well as thickness. Zhao and Zhang [8] proposed a uniform salient and discriminative descriptor learning method (SDDLM) for palmprint recognition in different scenarios, such as contact-based, contactless, and multispectral palmprint recognition.

Currently, deep learning model has been successfully used in many computer vision tasks, such as fingerprint recognition [31], face recognition [32], and iris recognition [33]. Deep learning-based methods have also been introduced to recognize palmprint [34]. Zhao et al. [35] trained the parameters of a deep belief net for palmprint recognition and obtained higher accuracy compared with traditional recognition methods. Shao et al. [36] generated fake images for an unlabeled palmprint dataset using a labeled dataset for cross-domain palmprint recognition. Genovese et al. [37] proposed a novel method for deep discriminative feature extraction based on Gabor responses and principal component analysis (PCA) called PalmNet. Based on convolutional neural networks (CNN), Zhao and Zhang [38] extracted high-level discriminative features and proposed deep discriminative representation (DDR) method for palmprint recognition. Shao et al. [39] concatenated the feature vectors obtained by several base networks to improve palmprint identification and verification accuracy, and proposed deep ensemble hashing method. Izadpanahkakkh et al. [40] extracted ROI and discriminative features for palmprint recognition using a neural network and transfer learning fusion method.

### B. Deep Hashing Network

DHN is mainly based on CNN and hashing algorithms to improve the efficiency of model. Due to the storage and retrieval efficiency, Zhu et al. [41] firstly proposed supervised hashing to approximate nearest neighbor search for large-scale multimedia retrieval, which improved the quality of hash coding by exploiting the semantic similarity on data pairs. Cao et al. [42] proposed Deep Visual-Semantic Quantization (DVSQ) to learn deep quantization models from labeled image data as well as the semantic information underlying general text domains. This framework learned deep visual-semantic embedding using carefully designed hybrid networks and loss functions. Lai et al. [43] proposed a deep architecture for supervised hashing to map images into binary codes. They designed a triplet ranking loss to produce the effective intermediate image features. Zhao et al. [44] proposed deep semantic ranking based hashing (DSRH) to preserve multilevel semantic similarity between multi-label images. They adopted an effective scheme based on surrogate loss to solve the problem of non-smooth and multivariate ranking measures. In order to generate compact and bit-scalable hashing codes from raw images, Zhang et al. [45] proposed a supervised learning framework using deep convolutional neural network in an end-to-end fashion. Experimental results showed this framework outperformed state-of-the-arts on public benchmarks and achieved promising results for person re-identification in surveillance. Qiao et al. [46] proposed Deep Heterogeneous Hashing (DHH) method to obtain unified binary codes for face images and videos, which consists of three stages, i.e. feature learning, video modeling, and heterogeneous hashing. Li et al. [47] adopted softmax classification loss and an improved triplet loss to learn the hash code to be consistent with the high-dimensional features. Liu et al. [48] proposed deep self-taught hashing algorithm (DSTH) to generates pseudo labels and then learns the hash codes using deep models.

TABLE I
THE DETAILS OF ACQUISITION DEVICES

| Device | iPhone 6S | HUAWEI Mate8 | LG G4 | Galaxy Note5 | MI8 |
|---|---|---|---|---|---|
| Pixels | 12 million | 16 million | 8 million | 16 million | 12 million |
| Image size | 3264×2448 | 3456×4608 | 5312×2988 | 5312×2988 | 4032×3024 |
| Aperture | f/2.2 | f/2.0 | f/1.8 | f/1.9 | f/1.8 |
| Sensor | BSI CMOS | CMOS | CMOS | CMOS | CMOS |
| GPU | PowerVR GX6450 | Mali-T880 | Adreno418 | Mali-760MP8 | Adreno630 |

### C. Knowledge Distillation

KD is an effective technique for transferring information from a relatively complex deep model to a light model [49]. It has been widely used for model compression and acceleration to improve the performance of fast and light networks [50]. Hinton et al. [51] first distilled knowledge from an ensemble of pre-trained models to improve a small target net via high-temperature softmax training. Then, FitNet developed KD using the pre-trained wide and shallow teachers hint layers to assist thin and deep students by guided layers [52]. Li et al. [53] proposed a unified distillation framework to use a small clean dataset and label relations in knowledge graph to guide the distillation process. Yim et al. [54] proposed a method of transferring the distilled knowledge as the flow between two layers by computing the inner product between features, and the student model outperformed the original model that was trained from scratch. Recently, KD is also successfully used for pedestrian detection [55] and face recognition [56]. Chen et al. [57] proposed a teacher bounded loss and introduced adaptation layers to help student network to better learn from teacher distributions for object detection. Shu et al. [58] adopted adversarial training stratagem to help learn the student network, which integrated merits from both process-oriented learning and result-oriented learning. Malinin et al. [59] combined KD with ensemble learning and proposed ensemble distribution distillation to improved classification performance. Park et al. [60] proposed relational knowledge distillation (RKD) to transfer relations of data examples based on distance-wise and angle-wise distillation losses. Wei et al. [61] quantized a large network and then mimicked a quantized small network for object detection. The model improved the performance of a student network by transferring knowledge from a teacher network. A research similar to our work is knowledge distillation-based metric learning, which is adopted to improve the image embedding. Yu et al. [13] proposed a novel distillation loss, which compared the distance between the features of two samples in the teacher network with the distance of the same two samples in the student network. Inspired by this idea, our proposed DDH is aimed to constrain the maximum distance or minimum distance of genuine and imposter matchings between teacher and student networks. In addition, hash coding is further adopted to improve the effectiveness of feature matching.

### III. XI'AN JIAOTONG UNIVERSITY UNCONSTRAINED PALMPRINT (XJTU-UP) DATABASE

#### A. Main Characteristics

XJTU-UP database is collected in unconstrained environments, which reduces the constraints of acquisition and improves the convenience of recognition system compared with other public palmprint databases. First of all, the collection devices are 5 commonly used smartphones launched in recent years, i.e. iPhone 6S, HUAWEI Mate8, LG G4, Samsung Galaxy Note5, and MI8. The details related to the acquisition devices are shown in Table I.

Secondly, during image acquisition, the volunteers took the smart phones by themselves and captured images indoors without other auxiliary equipment. They chose the angles and backgrounds of shooting as they wish, as long as the entire palm was captured. In order to increase the diversity of samples, the postures of palms and backgrounds in each image are continuously changed as much as possible. Two kinds of illumination are adopted, one is the natural illumination in the room and the other is the flash light of mobile phone. A total of 100 volunteers, 19 to 35 years old, provided their hand images. Using different mobile phones, each volunteer was asked to capture about 10 images of left or right hand under different illuminations. Therefore, for each volunteer, there are at least 5 (devices) × 2 (illuminations) × 2 (hands) × 10 (times)= 200 images with a total of 5 (devices) × 2 (illuminations) = 10 sub-datasets. All of the images are stored according to the original format of the phones, and the size of each RGB image is shown in Table I. Some typical palmprint images in XJTU-UP database are shown in Fig. 2.

TABLE II: COMPARISONS OF RELATED POPULAR PUBLIC PALMPRINT DATABASES

| Name | Type | Time | Main characteristics | Number of images | Number of hands | Domain | Manual key points |
|---|---|---|---|---|---|---|---|
| PolyU contacted[15] | Contacted | 2003 | Constrained | 7752 | 386 | 1 | No |
| PolyU multispectral[9] | Contacted | 2010 | Constrained | 24,000 | 500 | 4 | No |
| CASIA[17] | Contactless | 2005 | Constrained | 5502 | 624 | 1 | No |
| IIT-D [18] | Contactless | 2006 | Constrained | 2601 | 460 | 1 | No |
| Tongji [14] | Contactless | 2017 | Constrained | 12,000 | 600 | 1 | No |
| DPDS100 [62] | Contactless | 2011 | Constrained | 1000 | 100 | 1 | No |
| de Santos Sierra et. al.[63] | 1 mobile phone | 2011 | Uncomplicated | 30 per people | - | 1 | No |
| Choraś and Kozik [20] | 1 mobile phone | 2011 | Unconstrained | 252 | 84 | 1 | No |
| PRADD [21] | 2 mobile phones and 1 digital camera | 2012 | Unconstrained | 12,000 | 200 | 6 | No |
| Tiwari, Hwang and Gupta *et al.* [64] | 1 mobile phone | 2016 | Constrained | 186 videos | 62 | 1 | No |
| NUIG_Palm1 [22] | 5 mobile phones | 2017 | Unconstrained | 1616 | 81 | 10 | 5/image |
| NTU-PI-v1 [19] | Internet | 2019 | Uncooperative | 7781 | 2035 | 1 | No |
| NTU-CP-v1 [19] | Internet | 2019 | Uncooperative | 2478 | 655 | 1 | No |
| XJTU-UP (ours) | 5 mobile phones | 2020 | Unconstrained | > 20,000 | 200 | 10 | 14/image |

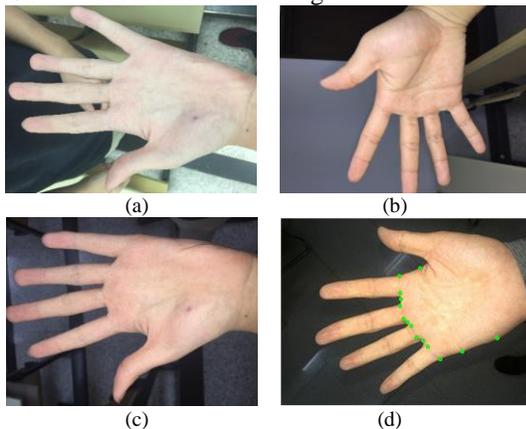

**Fig. 2.** Some typical examples in XJTU-UP database. (a), (b), and (c) are original images, and (d) is the image with 14 key points manually marked.

According to the acquisition devices and illuminations, 10 sub-datasets are named as IN (iPhone 6s under Natural illumination), IF (iPhone 6s under Flash illumination), HN (HUAWEI Mate8 under Natural illumination), HF (HUAWEI Mate8 under Flash illumination), LN (LG G4 under Natural illumination), LF (LG G4 under Flash illumination) SN (Samsung Galaxy Note5 under Natural illumination), SF (Samsung Galaxy Note5 under Flash illumination), MN (MI8 under Natural illumination), and MF (MI8 under Flash illumination).

In Table II, the currently published palmprint databases are summarized. The current researches are mainly based on palmprint databases collected in constrained environments, which are less affected by noises but relatively inconvenient. However, images captured on mobile phones in unconstrained environments are more flexible and convenient in practical applications, and correspondingly more complicated and difficult to identify. As can be seen from Table II, compared with XJTU-UP database, other databases are far less diverse. NUIG_Palm1 and XJTU-UP are most similar, which are collected from 5 different mobile phones in unconstrained environments and two illuminations. However, the number of samples in XJTU-UP is much larger than it, and the diversity of images is also much larger. To the best of our knowledge, XJTU-UP is the largest palmprint database currently open collected using mobile phones in unconstrained environments. Based on XJTU-UP, a lot of tasks can be carried out, such as stable ROI extraction, efficient and accurate palmprint identification, palmprint verification, recognition for open-set, and so on.

### B. ROI Extraction

In the palmprint recognition, stable ROI has a great influence on the accuracy of subsequent recognition algorithm. Typically, palmprint ROI extraction is based primarily on the contours and valleys of palm. After finding the key points, the central area of palm can be extracted as ROI based on distance [65], ratio [66], and angle [67]. However, in XJTU-UP, due to the complex background, it is difficult to extract the palm and its contours, especially under natural illuminations. In order to extract stable ROIs, 14 key points on the palm outline are manually marked in every image, including 3 valley points between fingers, 8 points at the bottom of fingers, and 3 points on either side of palm, as shown in Fig. 2 (d). The coordinates of these markers are also open to the academic community for research on ROI extraction. In this paper, we focus on the algorithm of efficient palmprint identification and verification, so ROIs manually extracted by the marks above are adopted, which can be seen as an example of research on ROI extraction on our XJTU-UP database and provide some convenience and inspirations for subsequent researchers.

As shown in Fig. 3, the valley between the index finger and the middle finger, $P_3$, and the valley between the ring finger and the little finger, $P_9$, are selected to determine the direction of ROI. The edge points on both palm sides, $P_0$ and $P_{12}$, which are less affected by the postures of hand, are selected to determine the length of ROI. The valley between the middle finger and the ring finger, $P_6$, is used to determine the center point, $P_o$, of ROI. Suppose the distance between $P_0$ and $P_{12}$ is $L$, therefore the side length of ROI is set to $\frac{3}{5}L$, and the distance between $P_o$ and $P_6$ is $\frac{2}{5}L$.

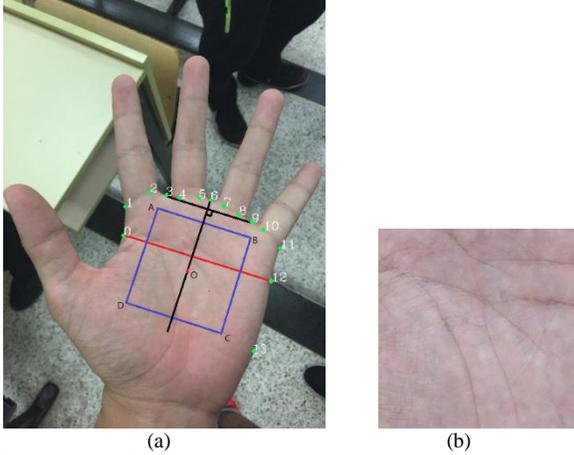

(a)                              (b)

**Fig. 3.** The details of ROI extraction. (a) is the schematic of ROI extraction and (b) is the ROI extracted. The green marked points are the 14 key points, the black line is the established coordinate system, the red line determines the side length of ROI, and the blue rectangle is the extracted ROI.

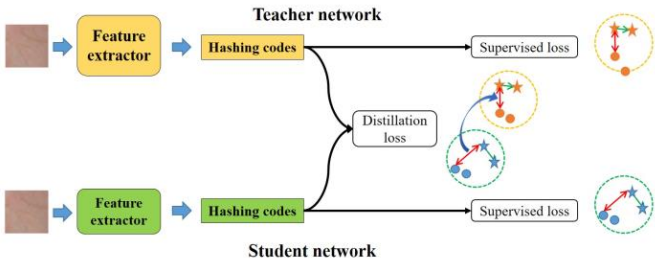

**Fig. 4.** The schematic of DDH proposed, which consists of a teacher network guiding to learn student network. DHN is adopted as backbone network, which converts palmprint images into binary codes. The teacher network is based on VGG-16, whose weights are firstly trained by supervised DHN loss and then fixed during training student network. Under the guidance of teacher, the student network, which has three fully connected layers and two convolutional layers, is trained by supervised DHN loss and distillation loss. DHN loss performs optimization on features and distillation loss performs optimization on the distances of feature matching pairs.

## IV. METHOD

DDH is proposed as a baseline for efficient palmprint recognition, which combines the advantages of both DHN and KD, and can improve the accuracy and efficiency. The schematic of DDH is shown in Fig. 4. Two DHNs are adopted as feature extractors, and the heavy one is used as teacher network and the light one is used as student network. Supervised loss is minimized to train DHNs on the features, and distillation loss is adopted to transfer the knowledge from teacher to help learn student network on the feature matching pairs.

### A. Deep Hashing Network

DHN is adopted as feature extractor to convert palmprint images into 128-bit binary codes. The storage space of binary codes is greatly compressed compared to the original image so that the storage efficiency is improved. In addition, a simple XOR operation can be used to obtain the Hamming distance between codes to improve the efficiency of matching. In this paper, DHN based on VGG-16 is adopted as the backbone of teacher network for palmprint recognition. VGG-16 has 5 batches of convolutional layers and 3 fully connected layers [68]. Thanks to the fine-tuning in transfer learning [69], the weights of pre-trained batch-1 to batch-4 on ImageNet are fine-tuned, as shown in Fig. 5. The batch-5 and fully connected layers are trained with actual palmprint images. In fact, DHN transforms the *softmax* layer of VGG-16 into a coding layer, where *sign* function is used as activation function. The optimization goal consists of two parts: hashing loss and quantization loss.

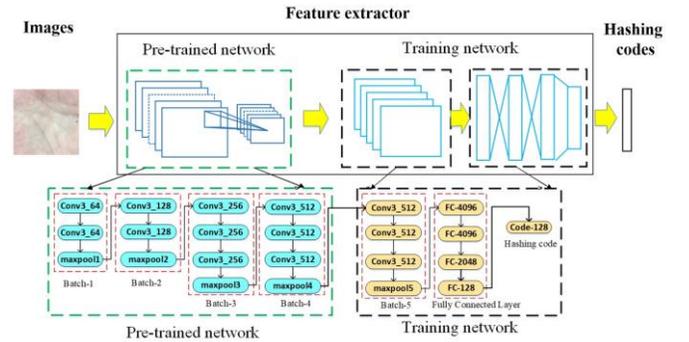

**Fig. 5.** The schematic diagram of DHN. The palmprint images are inputted to feature extractor to obtain binary codes. VGG-16 is adopted as backbone and the weights of batch-1 to batch-4 of pre-trained on ImageNet are fine-tuned directly. The batch-5 and fully connected layers are trained with actual palmprint images.

Hashing loss: hashing loss is based on contrastive loss. Suppose that for two images $i$ and $j$, the features extracted by DHN are denoted as $h_i$ and $h_j$, respectively, and the hashing loss between them is defined as:

$$L_{h_{i,j}} = \frac{1}{2}S_{i,j}D(h_i, h_j) + \frac{1}{2}(1 - S_{i,j})max(t - D(h_i, h_j), 0), \quad (1)$$

where $S_{i,j}$ represents their pair label. When images $i$ and $j$ form genuine matching, $S_{i,j} = 1$, otherwise, $S_{i,j} = 0$. $t$ is a margin threshold and set to 180, like [7].

Quantization loss: the quantization loss is mainly due to the fact that the feature is directly converted to binary code by *sign* function. It is defined as:

$$L_{q_i} = \|h_i - b_i\|_2 = \||h_i| - 1\|_2, \qquad (2)$$

where $b_i$ is the code of image $i$. Assuming there are a total of $N$ images, the optimization goal of DHN is

$$L_{DHN} = \sum_{i=1}^{N} \sum_{j=i+1}^{N} L_{h_{i,j}} + w \times \sum_{i=1}^{N} L_{q_i} = L_h + wL_q, \quad (3)$$

where $w$ is used to balance the weight between hashing loss and quantization loss.

### B. Knowledge Distillation for DHN

The above DHN-based palmprint recognition can obtain potential results. However, the model based on VGG-16 is relatively complex, which leads to long training time and low recognition efficiency. In order to solve this problem, a KD model is adopted here to realize efficient palmprint recognition under a light network. KD model includes a teacher network and a student network. In general, the teacher network is a complex network with strong feature extraction capabilities to ensure to extract distinguishable features. The student network adopts a light network with limited recognition capabilities but high recognition efficiency. In this paper, the above-mentioned VGG-16-based DHN is adopted as teacher network. A light network consisting of only three fully connected layers and two convolutional layers is adopted as student network. The parameters of student network are shown in Fig. 6.

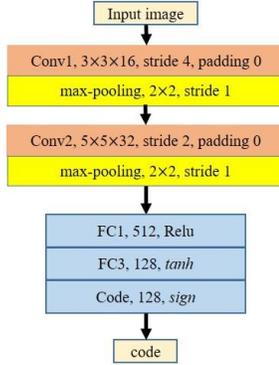

**Fig. 6.** The parameters of student network. "Conv" means convolutional network, "FC" means fully connected layer, and "Code" represents coding layer. The configuration of each layer is shown in the box. Activation functions are also demonstrated in each box.

For classification task, KD is aimed to encourage the student network to make a similar prediction as the teacher network. According to Hinton's method [51], teacher network will provide a soft label for student network using a temperature, $T$,

$$q_i^T = \frac{exp(z_i/T)}{\sum_i exp(z_i/T)}, \qquad (4)$$

where $z_i$ is the output of the $i$th neuron in last layer and $q_i$ is the probability that the input image belongs to the $i$th category. Through (4), the knowledge is transferred from teacher network to student network to improve its performance.

Similarly, for knowledge distillation for DHN, we would like the features extracted by the student network and teacher network as similar as possible. Naturally, the Euclidean distance between the features can be adopted as optimal object directly, denoted as direct quantization loss. Suppose for image $i$, the feature obtained by teacher network is $b_i^T$ and the feature obtained by student network is $b_i^S$, then the direct quantization loss is

$$L_{dir_i} = \|b_i^T - b_i^S\|_2. \qquad (5)$$

In [13], Yu *et al.* proposed relative teacher and enforced the student network to learn the distance of any two feature matching pairs extracted by teacher network. It is achieved by minimizing the loss:

$$L_{rela} = \|d_{ij}^T - d_{ij}^S\|_2, \qquad (6)$$

$$d_{ij}^T = \|b_i^T - b_j^T\|_2, \qquad (7)$$

$$d_{ij}^S = \|b_i^S - b_j^S\|_2, \qquad (8)$$

where $b_i^T$ and $b_j^T$ are features extracted by teacher network, and $b_i^S$ and $b_j^S$ are features extracted by student network of images $i$ and $j$.

Compared with (5), (6) can better optimize the differences between the features of teacher and student networks. Inspired by this idea, a new distillation loss is constructed to focus on hard sample mining, which optimizes the difference of maximum distances of genuine matchings and minimum distances of imposter matchings between the teacher and student networks. Generally, the teacher network is deeper and wider with stronger generalization and discrimination ability. Therefore, the teacher network knows exactly the limits of the distances between feature matchings. For DHN, the distance of genuine matchings is optimized to be as small as possible and the distance of imposter matchings is enforced to be as large as possible. Therefore, if the maximum distance of genuine matching of student network is smaller than the minimum distance of imposter matching, and the minimum distance of imposter matching of student network is bigger than the maximum distance of imposter matching of teacher network, the performance of student will be improved greatly. The new distillation loss is donated as

$$L_{hard} = \begin{cases} \max_{i,j}(d_{ij}^S) - \min_{i,j}(d_{ij}^T) & \text{for genuines} \\ \max_{i,j}(d_{ij}^T) - \min_{i,j}(d_{ij}^S) & \text{for imposters} \end{cases}. (9)$$

Therefore, the student network is trained by simultaneously considering the standard DHN loss, $L_{DHN}$, and the distillation loss imposed by the teacher, according to:

$$min\, L = L_{DHN} + \alpha L_{rela} + \beta L_{hard}, \qquad (10)$$

where $\alpha$ and $\beta$ are trade-off parameters between different losses.

## V. EXPERIMENTS AND RESULTS

### A. Databases

In addition to XJTU-UP database, another two benchmarks are also adopted to perform experiments to evaluate DDH, and the details are as follow:

#### 1) PolyU multispectral palmprint database

PolyU multispectral palmprint database contains four different spectral bands, *i.e.* blue, red, green and near-infrared (NIR) [9]. Under each spectral band, there are 6,000 grayscale images from 500 different palms of 250 individuals, including 195 males and 55 females. For each palm, 12 images were collected in two sessions, 6 images in each session. The palmprint images are oriented and cropped to form the ROIs with the size of 128×128 pixels. According to different spectral bands, this database can be used as four sub-datasets, recorded as Blue, Red, Green, and NIR. Fig. 7 shows some typical ROI samples of them.

**2) Tongji contactless palmprint database**

Tongji palmprint database was collected in 2017, which is one of the biggest contactless palmprint databases that time [14]. It contains 12,000 images captured from 300 individuals, including 192 males and 108 females. For each category, there are 20 palm images. During acquisition, the volunteers can move their hands up and down freely within the enclosed space. However, a ring white LED light source was adopted and the background was still pure black, which reduces the difficulty of palmprint recognition. Some typical examples are shown in Fig. 8 and ROIs with the size of 128×128 pixels are extracted based on [14].

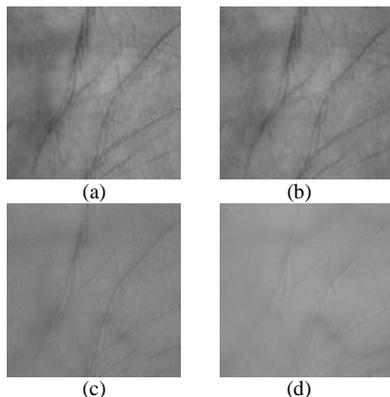

**Fig. 7.** Some typical ROI samples of PolyU multispectral palmprint database. (a) is in Blue, (b) is in Green, (c) is in Red, and (d) is in NIR.

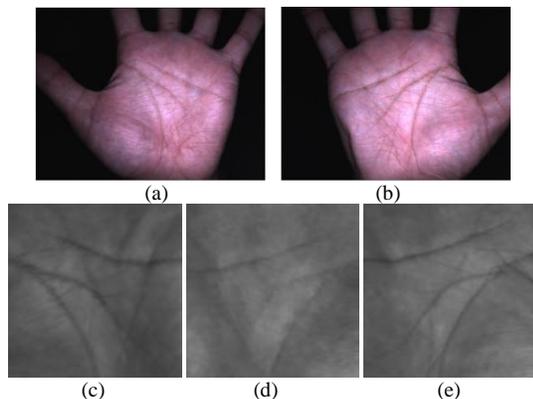

**Fig. 8.** Some typical samples of Tongji palmprint database. (a) and (b) are original images; (c), (d), and (e) are ROIs.

*B. Implementation Details*

During experiments, the teacher network is firstly trained by supervised DHN loss, *i.e.* (3). Then the teacher weights are fixed to provide guidance to student through (9). Finally, the student network is trained by (10). During testing, palmprint identification and verification are performed on different datasets. For each category, the half of palmprint images are adopted as training set and the remaining half are used as test set to evaluate the performance of DDH. For palmprint identification, after extracting features, every testing image is matched with all of the training images to find the most similar one to it. If they are from the same subject, it shows that they are genuine matching and the identification is successful. Then the accuracy can be calculated. For palmprint verification, the testing images are also matched with the training images to calculate the Hamming distance. Then, the False Acceptance Rate (FAR), False Rejection Rate (FRR), and EER can be calculated. The experiments are implemented using TensorFlow framework on a NVIDIA GPU GTX1080 with 8G memory power. The base learning rate is set to 0.001; and the Adam Optimizer and Stochastic Gradient Descent (SGD) methods are adopted.

*C. Palmprint Identification*

**Performance on XJTU-UP database**: XJTU-UP database consists of 10 sub-datasets. For each subject, 5 images are used for training and the reaming 5 images are used for testing. The results of identification accuracy are listed in Table III. From the results, the average accuracy of teacher network is 98.68% and the maximum accuracy can be up to 99.27%, which indicates the feasibility of our XJTU-UP database. The performance of student network is relatively bad, and the accuracy is 80.76%. While our DDH can improve the average accuracy to 85.94%. In IN, the accuracy is improved by up to 11.37%, which demonstrates the effectiveness of DDH.

TABLE III
ACCURACY (%) OF PALMPRINT IDENTIFICATION FOR XJTU-UP DATABASE

| Databases | Student | DDH | Teacher |
|-----------|---------|-------|---------|
| IF | 89.49 | 90.10 | 98.78 |
| IN | 75.43 | 86.80 | 97.56 |
| HF | 83.64 | 87.58 | 98.38 |
| HN | 70.10 | 77.90 | 99.10 |
| LF | 84.97 | 90.47 | 99.27 |
| LN | 78.80 | 79.70 | 97.70 |
| MF | 85.15 | 88.45 | 98.97 |
| MN | 77.92 | 84.38 | 98.96 |
| SF | 84.38 | 90.42 | 99.27 |
| SN | 77.74 | 83.59 | 98.77 |
| Average | 80.76 | 85.94 | 98.68 |

**Performance on PolyU multispectral palmprint database**: for PolyU multispectral palmprint database, there are 4 sub-datasets consisting of 6,000 palmprint images. For each subject, 6 images are used for training and the remaining 6 images are used for testing. Table IV shows the accuracy of palmprint identification. From the results, the best performance of teacher network is obtained in Red, where the accuracy of teacher is 100%. In Blue, DDH obtains the biggest improvement in performance, where the accuracy is improved by up to 7.63% compared with the student. As a result, at the guidance of teacher network, the accuracy of student network is improved by 5.22%.

**Performance on Tongji contactless palmprint database**: in Tongji palmprint database, there are 20 samples for each subjects, where 10 samples are used for training and the remaining 10 images are used for testing. The results are list in Table V. The accuracy of teacher is 100%, and DDH improves the accuracy from 97.85% of student to 98.92%.

TABLE IV
ACCURACY (%) OF PALMPRINT IDENTIFICATION FOR POLYU MULTISPECTRAL
PALMPRINT DATABASE

| Databases | Student | DDH | Teacher |
|-----------|---------|-------|---------|
| Blue | 87.87 | 95.50 | 99.97 |
| Green | 86.40 | 91.93 | 99.90 |
| Red | 93.10 | 96.13 | 100.00 |
| NIR | 91.27 | 95.97 | 99.83 |
| Average | 89.66 | 94.88 | 99.93 |

TABLE V
ACCURACY (%) OF PALMPRINT IDENTIFICATION FOR XJTU-UP DATABASE

| Databases | Student | DDH | Teacher |
|-----------|---------|-------|---------|
| Tongji | 97.85 | 98.92 | 100.00 |

### D. Palmprint Verification

**Performance on XJTU-UP database**: for each sub-dataset, genuine matchings and imposter matchings are obtained. The EERs of palmprint verification are shown in Table VI. The teachers obtain the best performances, where the EERs are lower than 0.47%. Compared with others, the results of HN is a little bad, which may be because of its complex illuminations. The average EER of teacher is 0.29%, and the average EER is reduced from 4.12% of student to 2.84% by DDH. The biggest improvement is obtained in HN, where EER is reduced by 3.11%.

TABLE VI
EER (%) OF PALMPRINT VERIFICATION FOR XJTU-UP DATABASE

| Databases | Student | DDH | Teacher |
|-----------|---------|------|---------|
| IF | 1.34 | 1.12 | 0.24 |
| IN | 4.86 | 2.52 | 0.37 |
| HF | 3.51 | 2.68 | 0.27 |
| HN | 8.17 | 5.06 | 0.20 |
| LF | 3.00 | 1.66 | 0.22 |
| LN | 5.27 | 5.13 | 0.47 |
| MF | 3.23 | 2.41 | 0.21 |
| MN | 4.04 | 2.67 | 0.30 |
| SF | 2.73 | 1.77 | 0.24 |
| SN | 5.08 | 3.36 | 0.36 |
| Average | 4.12 | 2.84 | 0.29 |

TABLE VII
EER (%) OF PALMPRINT VERIFICATION FOR POLYU MULTISPECTRAL
PALMPRINT DATABASE

| Databases | Student | DDH | Teacher |
|-----------|---------|------|---------|
| Blue | 2.52 | 1.02 | 0.0012 |
| Green | 2.79 | 1.73 | 0.0055 |
| Red | 1.77 | 1.08 | 0.0061 |
| NIR | 2.31 | 0.95 | 0.0054 |
| Average | 2.35 | 1.20 | 0.0046 |

**Performance on PolyU multispectral palmprint database**: for each sub-dataset, there are 18,000 genuine matchings and 898,200 imposter matchings. The results are shown in Table

VII. The average EER of teacher network is as low as 0.0046%, which shows its effectiveness. The average EER of student is 2.35%, which is reduced to 1.20% by DDH.

**Performance on Tongji contactless palmprint database**: in Tongji palmprint database, there are 12,000 images belonging to 600 categories. 60,000 genuine matchings and 35,940,000 imposter matchings are obtained. The results are shown in Table VIII. The EER of teacher is 0.013%, and the EER of DDH is 2.53%, which is much lower than 3.89% obtained by student.

TABLE VIII
EER (%) OF PALMPRINT VERIFICATION FOR TONGJI CONTACTLESS PALMPRINT
DATABASE

| Databases | Student | DDH | Teacher |
|-----------|---------|------|---------|
| Tongji | 3.89 | 2.53 | 0.013 |

## VI. EVALUATION AND ANALYSES

### A. Result Analysis

Palmprint identification and verification are performed on one unconstrained and two constrained databases, *i.e.* XJTU-UP, PolyU multispectral, and Tongji contactless databases, using teacher network, student network, and DDH. From the results, the teacher can obtain the optimal results on all of the experiments, but the model is complex and has a lot of parameters which costs much time for training. The student is lighter and can save training resources, but the accuracy is relatively low. Our proposed DDH can obtain better performance using the same network as the student. The results indicate the effectiveness of DDH.

In addition, compared with XJTU-UP database, the constrained databases, PolyU multispectral and Tongji contactless databases, can achieve better performance both on accuracy and EERs. Constrained databases are collected from enclosed space with additional lights, which are less effected by noises and outside illuminations. Though they are easier to identify, their acquisition limits the convenience of palmprint recognition. XJTU-UP database is collected in an unconstrained manner by smart phones, which is more suitable for practical applications. From the results above, our DDH algorithm can obtain relatively satisfactory results on XJTU-UP database, which indicates the feasibility of our unconstrained database and the potential of palmprint recognition to be used as an efficient biometrics for deployment on consumer devices.

### B. Ablation Study

1) The roles of different losses

In this paper, different distillation losses are added to DHN to improve the accuracy of student network., Here we conducted several experiments on IF, IN, HF, and HN databases to verify the roles that different distillation losses play. The results are shown Tables IX and X. $L_{dir}$ is adopted to constrain the distance of the same sample for teacher and student networks. $L_{rela}$ is adopted to reduce the relative distance of feature matching pairs extracted by teacher and student networks. DHN is aimed to reduce the distance of genuine matchings and increase the distance of imposter matchings, which is also performed on feature matching pairs. So $L_{rela}$ is more effective than $L_{dir}$, as shown in Tables IX and X. Further, based on $L_{rela}$, $L_{hard}$ is aimed to improve the performance of student by constraining the relationship between the maximum or minimum distances of the feature matching pairs extracted by teacher and student. Teachers know the limits of the maximum or minimum distance between feature matching pairs, so it is more effective, which can also be indicated by the results in Tables IX and X.

TABLE IX
ABLATION RESULTS (ACCURACY, %) OF PALMPRINT IDENTIFICATION FOR XJTU-UP DATABASE

| Databases | $L_{DHN}$ | $L_{DHN}$ $+L_{dir}$ | $L_{DHN}$ $+L_{rela}$ | $L_{DHN}+L_{rela}$ $+L_{hard}$ |
|---|---|---|---|---|
| IF | 89.49 | 87.96 | 89.19 | 90.10 |
| IN | 75.43 | 82.03 | 84.87 | 86.80 |
| HF | 83.64 | 85.45 | 87.07 | 87.58 |
| HN | 70.10 | 70.50 | 75.70 | 77.90 |
| LF | 84.97 | 89.64 | 87.15 | 90.47 |
| LN | 78.80 | 76.40 | 75.20 | 79.70 |
| MF | 85.15 | 84.23 | 87.43 | 88.45 |
| MN | 77.92 | 80.10 | 82.08 | 84.38 |
| SF | 84.38 | 85.31 | 89.27 | 90.42 |
| SN | 77.74 | 76.01 | 80.62 | 83.59 |

TABLE X
ABLATION RESULTS (EER, %) OF PALMPRINT VERIFICATION FOR XJTU-UP DATABASE

| Databases | $L_{DHN}$ | $L_{DHN}$ $+L_{dir}$ | $L_{DHN}$ $+L_{rela}$ | $L_{DHN}+L_{rela}$ $+L_{hard}$ |
|---|---|---|---|---|
| IF | 1.34 | 1.47 | 1.34 | 1.12 |
| IN | 4.86 | 4.23 | 3.14 | 2.52 |
| HF | 3.51 | 3.53 | 2.85 | 2.68 |
| HN | 8.17 | 7.35 | 6.22 | 5.06 |
| LF | 3.00 | 2.80 | 2.04 | 1.66 |
| LN | 5.27 | 7.27 | 5.69 | 5.13 |
| MF | 3.23 | 3.44 | 2.79 | 2.41 |
| MN | 4.04 | 3.38 | 3.02 | 2.67 |
| SF | 2.73 | 2.48 | 1.79 | 1.77 |
| SN | 5.08 | 5.06 | 4.46 | 3.36 |

TABLE XI
ACCURACY (%) OF PALMPRINT IDENTIFICATION FOR DIFFERENT HYPERPARAMETERS

| Databases | IF | IN | HF | HN |
|---|---|---|---|---|
| $\alpha = 0.5, \beta = 0.8$ | 88.98 | 85.79 | 84.14 | 76.43 |
| $\alpha = 0.8, \beta = 0.8$ | 88.98 | 85.79 | 86.67 | 75.81 |
| $\alpha = 1, \beta = 0.8$ | **90.10** | **86.80** | **87.58** | **77.90** |
| $\alpha = 1.5, \beta = 0.8$ | 89.29 | 81.02 | 85.56 | 76.02 |
| $\alpha = 1, \beta = 0.1$ | 89.80 | 82.74 | 86.57 | 77.00 |
| $\alpha = 1, \beta = 0.5$ | 89.49 | 83.35 | 87.27 | 74.44 |
| $\alpha = 1, \beta = 1$ | 89.39 | 84.47 | 87.27 | 77.20 |
| $\alpha = 1, \beta = 2$ | 87.86 | 85.99 | 86.97 | 76.70 |

### 2) The effect of Hyperparameters

In this part, we conducted several experiments on IF, IN, HF, and HN databases to show the effects of main hyperparameters,

$\alpha$ and $\beta$, which are used to balance the weights of different losses, and the results are in Tables XI and XII. From the results, the best performance can be obtained when $\alpha = 1$ and $\beta = 0.8$.

TABLE XII
EER (%) OF PALMPRINT VERIFICATION FOR DIFFERENT HYPERPARAMETERS

| Databases | IF | IN | HF | HN |
|---|---|---|---|---|
| $\alpha = 0.5, \beta = 0.8$ | 1.53 | 2.61 | 3.32 | 6.25 |
| $\alpha = 0.8, \beta = 0.8$ | 1.59 | **2.52** | 2.83 | 6.05 |
| $\alpha = 1, \beta = 0.8$ | **1.12** | **2.52** | **2.68** | **5.06** |
| $\alpha = 1.5, \beta = 0.8$ | 1.29 | 3.72 | 3.04 | 5.53 |
| $\alpha = 1, \beta = 0.1$ | 1.58 | 3.52 | 2.84 | 5.10 |
| $\alpha = 1, \beta = 0.5$ | 1.20 | 3.10 | 3.12 | 6.33 |
| $\alpha = 1, \beta = 1$ | 1.66 | 3.30 | 2.88 | 5.54 |
| $\alpha = 1, \beta = 2$ | 1.48 | 3.40 | 2.93 | 5.65 |

### C. Comparisons with Other Models

For comparison, we present the results of some baseline methods on palmprint identification and verification using XJTU-UP database, as follow:

- **Alexnet** [70] is adopted for contactless palmprint verification of new-born babies and trained by *softmax* loss function.
- **BNN** [71] is an efficient algorithm to drastically reduce the memory size of networks by replacing most arithmetic operations with bit-wise operations. Here, BNN is adopted to compress the student network.
- **Hinton's** [51] distills knowledge from teacher network to improve the student network via high-temperature *softmax* training.
- **$L_{dir}$+ Hinton's** [12] is proposed by our previous work for efficient palmprint verification and classification, which introduces knowledge distillation in to DHN and adopts $L_{dir}$ as distillation loss.
- **PKT** [72] performs knowledge distillation by matching the probability distribution of features extracted by teacher and student networks, which is an efficient knowledge distillation method for embeddings and similar to DDH.
- **Relative teacher** [13] is similar to $L_{rela}$, which addresses knowledge distillation for embeddings through the distance of feature pairs.

Note that all of modules are implemented using similar hyperparameters with a slight difference in each model to reach the best performance, respectively. The results of comparisons are shown in Tables XIII and XIV. From the results, teacher obtain the optimal results on palmprint identification and verification. Though Hinton's was proved to achieve high accuracy on classification task, it seems that it cannot be succeed in extracting discriminative embeddings. In contrast, PKT and relative teacher can obtain better results. Alexnet obtains higher identification accuracy on HF and SF databases, but the performance of palmprint verification is not very satisfactory. However, as a result, our proposed DDH can outperform other compression and knowledge distillation algorithms to achieve the state-of-the-arts.

TABLE XIII
COMPARISON (ACCURACY, %) OF PALMPRINT IDENTIFICATION ON XJTU-UP DATABASE

| Methods | IF | IN | HF | HN | LF | LN | MF | MN | SF | SN |
|---|---|---|---|---|---|---|---|---|---|---|
| Student | 89.49 | 75.43 | 83.64 | 70.10 | 84.97 | 78.80 | 85.15 | 77.92 | 84.38 | 77.74 |
| Alexnet | 83.57 | 76.75 | **87.78** | 74.50 | 76.06 | 60.60 | 87.73 | 81.04 | **92.39** | 78.97 |
| BNN | 86.63 | 70.36 | 82.83 | 55.20 | 83.94 | 63.80 | 82.47 | 67.60 | 83.54 | 71.18 |
| Hinton's | 35.20 | 51.27 | 38.28 | 44.80 | 31.71 | 43.30 | 37.53 | 33.96 | 18.96 | 37.74 |
| $L_{dir}$+ Hinton's | 88.06 | 83.05 | 75.60 | 74.34 | 76.00 | 79.48 | 83.20 | 82.92 | 82.71 | 79.36 |
| $L_{dir}$ | 87.96 | 82.03 | 85.45 | 70.50 | 89.64 | 76.40 | 84.23 | 80.10 | 85.31 | 76.01 |
| PKT | 87.65 | 79.29 | 85.66 | 67.80 | 87.04 | 69.90 | 85.98 | 74.79 | 88.44 | 76.00 |
| Relative teacher | 89.19 | 84.87 | 87.07 | 75.70 | 87.15 | 75.20 | 87.43 | 82.08 | 89.27 | 80.62 |
| DDH | **90.10** | **86.80** | 87.58 | **77.90** | **90.47** | **79.70** | **88.45** | **84.38** | 90.42 | **83.59** |
| Teacher | 98.78 | 97.56 | 98.38 | 99.10 | 99.27 | 97.70 | 98.97 | 98.96 | 99.27 | 98.77 |

TABLE XIV
COMPARISON (EER, %) OF PALMPRINT VERIFICATION ON XJTU-UP DATABASE

| Methods | IF | IN | HF | HN | LF | LN | MF | MN | SF | SN |
|---|---|---|---|---|---|---|---|---|---|---|
| Student | 1.34 | 4.86 | 3.51 | 8.17 | 3.00 | 5.27 | 3.23 | 4.04 | 2.73 | 5.08 |
| Alexnet | 3.64 | 7.75 | 4.07 | 7.87 | 7.53 | 12.90 | 4.13 | 6.22 | 3.03 | 7.23 |
| BNN | 1.82 | 7.64 | 4.14 | 11.97 | 3.15 | 9.07 | 7.57 | 5.44 | 3.06 | 7.26 |
| Hinton's | 21.64 | 12.09 | 17.32 | 17.92 | 20.94 | 16.52 | 23.73 | 19.74 | 25.23 | 19.23 |
| $L_{dir}$+ Hinton's | 1.41 | 3.61 | 3.67 | 6.35 | 2.66 | 5.81 | 2.55 | 2.76 | 2.12 | 4.82 |
| $L_{dir}$ | 1.47 | 4.23 | 3.53 | 7.35 | 2.80 | 7.27 | 3.44 | 3.38 | 2.48 | 5.06 |
| PKT | 1.31 | 4.94 | 3.67 | 8.58 | 2.53 | 7.42 | 3.43 | 4.55 | 2.14 | 5.56 |
| Relative teacher | 1.34 | 3.14 | 2.85 | 6.22 | 2.04 | 5.69 | 2.79 | 3.02 | 1.79 | 4.46 |
| DDH | **1.12** | **2.52** | **2.68** | **5.06** | **1.66** | **5.13** | **2.41** | **2.67** | **1.77** | **3.36** |
| Teacher | 0.24 | 0.37 | 0.27 | 0.20 | 0.22 | 0.47 | 0.21 | 0.30 | 0.24 | 0.36 |

## D. Computational Time Cost Analysis:

To evaluate the computational complexity of the proposed methods, we calculate the computational time cost and compare it with other methods, as shown in Table XV. The model size, parameters, average feature extraction time, and feature matching time are obtained. The size and complexity of teacher network are the largest, which is much larger than of other models including student network. At the same time, from the feature extraction time, the teacher network is the slowest, about 4 times the student network. BNN has the smallest memory space and the fastest processing speed. However, combined with the above analysis, the accuracy of BNN declines relatively seriously. For DDH, although it is necessary to obtain the trained teacher network before training the student network, the performance has been significantly improved. This is very important for the palmprint recognition system on the consumer device, because the complex model cannot work well with limited computing resources.

TABLE XV
COMPUTATIONAL COST OF DIFFERENT MODELS

| | Teacher network | Student network | DDH | Alexnet | BNN |
|---|---|---|---|---|---|
| Model size | 93 M | 1.19 M | 1.19 M | 3.5 M | 0.30 M |
| Total parameters | 156 M | 12.04 M | 12.04 M | 57.38M | 12.04 M |
| Feature extraction time | 12.30 ms | 3.33 ms | 3.32 ms | 5.78 ms | 2.86 ms |
| Feature matching time | 0.011ms | 0.012 ms | 0.011 ms | 0.142 ms | 0.011 ms |
| Iterations | 10,000 | 10,000 | 10,000 | 10,000 | 10,000 |

## VII. CONCLUSION

In this paper, DDH is proposed for efficient palmprint identification and verification. Firstly, based on DHN, a heavy network is trained as teacher network and a light network is trained as student network. The similarity of palmprint images can be obtained easily by calculating the Hamming distance between hashing codes, which can improve the efficiencies of feature matching. Then, a novel distillation loss function is constructed to help to learn student network at the guidance of teacher network. Finally, palmprint recognition can be performed using more efficient feature extraction and matching. Further, a new unconstrained palmprint database is built using 5 smart phones, called XJTU-UP database. Compared to other public databases, it has fewer restrictions on acquisition, so it is more difficult to recognition, but at the same time more flexible and convenient. There are 14 key points manually labeled on each image for ROI extraction, which are publicly available to the research community. Extensive experiments are conducted both on unconstrained XJTU-UP database and other two constrained databases, PolyU multispectral and Tongji contactless databases. From the results, using DDH, the accuracy of palmprint identification can be increased by up to 11.37%, and the EER of palmprint verification can be reduced by up to 3.11%. The experimental results show that our proposed models can outperform other baseline models, and the database we proposed has many advantages and can be successfully used in the research of palmprint recognition.